\documentclass[10pt,twocolumn,letterpaper]{article}

\usepackage{cvpr}
\usepackage{times}
\usepackage{epsfig}
\usepackage{graphicx}
\usepackage{amsmath}
\usepackage{amssymb}

\usepackage{color}

\usepackage{booktabs}
\usepackage{multirow}
\usepackage{array}
\usepackage{xcolor}
\usepackage{subcaption}
\usepackage{graphicx}
\usepackage{hhline}
\usepackage{makecell}
\usepackage{colortbl}
\usepackage[normalem]{ulem}
\usepackage{soul}
\usepackage{url}
\definecolor{Gray}{rgb}{0.7,0.7,0.7}
\usepackage[square,comma,numbers,sort]{natbib}

\renewcommand{\hl}[1]{\underline{\textbf{#1}}}
\usepackage[
	pagebackref=true,
	breaklinks=true,
	letterpaper=true,
	colorlinks,
	bookmarks=false,
	citecolor=red,
	linkcolor=blue]{hyperref}

\cvprfinalcopy 

\def\cvprPaperID{192} 

\ifcvprfinal\fi
\begin{document}

\title{Training Region-based Object Detectors with Online Hard Example Mining}
\author{
Abhinav Shrivastava$^1$ \qquad
Abhinav Gupta$^1$ \qquad
Ross Girshick$^2$ \\
$^1$Carnegie Mellon University \qquad
$^2$Facebook AI Research\\
{\tt\small \{ashrivas,abhinavg\}@cs.cmu.edu} \qquad
{\tt\small rbg@fb.com}
}
\maketitle

\begin{abstract}
The field of object detection has made significant advances riding on the wave of region-based ConvNets, but their training procedure still includes many heuristics and hyperparameters that are costly to tune. We present a simple yet surprisingly effective online hard example mining (OHEM) algorithm for training region-based ConvNet detectors. Our motivation is the same as it has always been -- detection datasets contain an overwhelming number of easy examples and a small number of hard examples. Automatic selection of these hard examples can make training more effective and efficient. OHEM is a simple and intuitive algorithm that eliminates several heuristics and hyperparameters in common use. But more importantly, it yields consistent and significant boosts in detection performance on benchmarks like PASCAL VOC 2007 and 2012. Its effectiveness increases as datasets become larger and more difficult, as demonstrated by the results on the MS COCO dataset. Moreover, combined with complementary advances in the field, OHEM leads to state-of-the-art results of 78.9\% and 76.3\% mAP on PASCAL VOC 2007 and 2012 respectively.
\end{abstract}

\vspace{-0.15in}

\section{Introduction}\label{sec:intro}
Image classification and object detection are two fundamental computer vision tasks.
Object detectors are often trained through a \emph{reduction} that converts object detection into an image classification problem. This reduction introduces a new challenge that is not found in natural image classification tasks: the training set is distinguished by a large imbalance between the number of annotated objects and the number of \emph{background} examples (image regions not belonging to any object class of interest). In the case of sliding-window object detectors, such as the deformable parts model (DPM) \cite{dpm}, this imbalance may be as extreme as 100,000 background examples to every one object. The recent trend towards object-proposal-based detectors \cite{Uijlings13,rcnn} mitigates this issue to an extent, but the imbalance ratio may still be high (\eg, 70:1). This challenge opens space for learning techniques that cope with imbalance and yield faster training, higher accuracy, or both.

Unsurprisingly, this is not a new challenge and a standard solution, originally called \emph{bootstrapping} (and now often called \emph{hard negative mining}), has existed for at least 20 years. Bootstrapping was introduced in the work of Sung and Poggio \cite{sungThesis} in the mid-1990's (if not earlier) for training face detection models. Their key idea was to gradually grow, or \emph{bootstrap}, the set of background examples by selecting those examples for which the detector triggers a false alarm. This strategy leads to an iterative training algorithm that alternates between updating the detection model given the current set of examples, and then using the updated model to find new false positives to add to the bootstrapped training set. The process typically commences with a training set consisting of all object examples and a small, random set of background examples.

Bootstrapping has seen widespread use in the intervening decades of object detection research. Dalal and Triggs \cite{DalalTriggs} used it when training SVMs for pedestrian detection. Felzenszwalb \etal \cite{dpm} later proved that a form of bootstrapping for SVMs converges to the global optimal solution defined on the entire dataset. Their algorithm is often referred to as \emph{hard negative mining} and is frequently used when training SVMs for object detection \cite{Uijlings13,rcnn,SPPnet}. Bootstrapping was also successfully applied to a variety of other learning models, including shallow neural networks \cite{rowley1998neural} and boosted decision trees \cite{dollar}. Even modern detection methods based on deep convolutional neural networks (ConvNets) \cite{backprop,AlexNet}, such as R-CNN \cite{rcnn} and SPPnet \cite{SPPnet}, still employ SVMs trained with hard negative mining.

It may seem odd then that the current state-of-the-art object detectors, embodied by Fast R-CNN \cite{frcn} and its descendants \cite{ren2015faster}, do not use bootstrapping. The underlying reason is a technical difficulty brought on by the shift towards purely online learning algorithms, particularly in the context of deep ConvNets trained with stochastic gradient descent (SGD) on millions of examples. Bootstrapping, and its variants in the literature, rely on the aforementioned alternation template: (a) for some period of time a \emph{fixed} model is used to find new examples to add to the active training set; (b) then, for some period of time the model is trained on the \emph{fixed} active training set. Training deep ConvNet detectors with SGD typically requires hundreds of thousands of SGD steps and freezing the model for even a few iterations at a time would dramatically slow progress. What is needed, instead, is a purely online form of hard example selection.

In this paper, we propose a novel bootstrapping technique called \emph{online hard example mining}\footnote{We use the term hard \emph{example} mining, rather than hard \emph{negative} mining, because our method is applied in a multi-class setting to all classes, not just a ``negative'' class.} (OHEM) for training state-of-the-art detection models based on deep ConvNets. The algorithm is a simple modification to SGD in which training examples are sampled according to a non-uniform, non-stationary distribution that depends on the current loss of each example under consideration. The method takes advantage of detection-specific problem structure in which each SGD mini-batch consists of only one or two images, but \emph{thousands} of candidate examples. The candidate examples are subsampled according to a distribution that favors diverse, high loss instances. Gradient computation (backpropagation) is still efficient because it only uses a small subset of all candidates. We apply OHEM to the standard Fast R-CNN detection method and show three benefits compared to the baseline training algorithm:

\begin{itemize}
\item It removes the need for several heuristics and hyperparameters commonly used in region-based ConvNets.
\item It yields a consistent and significant boosts in mean average precision.
\item Its effectiveness increases as the training set becomes larger and more difficult, as demonstrated by results on the MS COCO dataset.
\end{itemize}

Moreover, the gains from OHEM are complementary to recent improvements in object detection, such as multi-scale testing \cite{SPPnet} and iterative bounding-box regression \cite{mrcnn}. Combined with these tricks, OHEM gives state-of-the-art results of \textbf{78.9\%} and \textbf{76.3\%} mAP on PASCAL VOC 2007 and 2012, respectively.


\begin{figure*}[!ht]
	\centering
	\includegraphics[width=\linewidth]{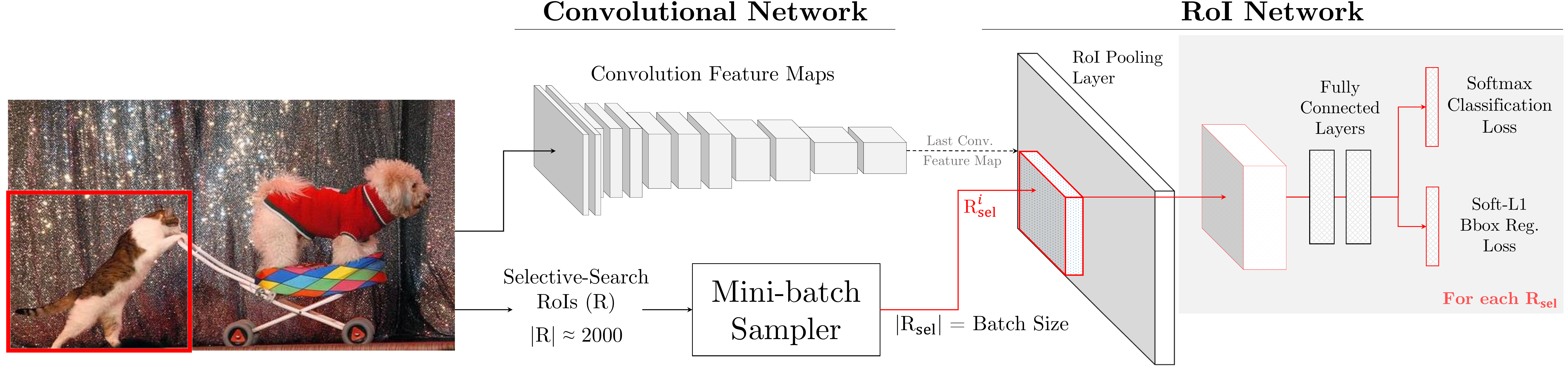}
  \caption[caption]{Architecture of the Fast R-CNN approach (see Section \protect\ref{sec:frcn-overview-sec} for details).}\vspace{-0.15in}
	\label{fig:frcn-overview}
\end{figure*}

\section{Related work}
Object detection is one of the oldest and most fundamental problems in computer vision. The idea of dataset \emph{bootstrapping} \cite{sungThesis,rowley1998neural}, typically called \emph{hard negative mining} in recent work \cite{dpm}, appears in the training of most successful object detectors \cite{rowley1998neural,dollar,DalalTriggs,esvm,midlevel,dpm,rcnn,SPPnet,mrcnn}. Many of these approaches use SVMs as the detection scoring function, even after training a deep convolutional neural network (ConvNet) \cite{backprop,AlexNet} for feature extraction. One notable exception is the Fast R-CNN detector \cite{frcn} and its descendants, such as Faster R-CNN \cite{ren2015faster}. Since these models do not use SVMs, and are trained purely online with SGD, existing hard example mining techniques cannot be immediately applied. This work addresses that problem by introducing an online hard example mining algorithm that improves optimization and detection accuracy. We briefly review hard example mining, modern ConvNet-based object detection, and relationships to concurrent works using hard example selection for training deep networks.

\vspace{-0.1in}
\paragraph{Hard example mining.}
There are two hard example mining algorithms in common use. The first is used when optimizing SVMs. In this case, the training algorithm maintains a working set of examples and alternates between training an SVM to convergence on the working set, and updating the working set by removing some examples and adding others according to a specific rule \cite{dpm}. The rule removes examples that are ``easy'' in the sense that they are correctly classified beyond the current model's margin. Conversely, the rule adds new examples that are hard in the sense that they violate the current model's margin. Applying this rule leads to the global SVM solution. Importantly, the working set is usually a small subset of the entire training set. 

The second method is used for non-SVMs and has been applied to a variety of models including shallow neural networks \cite{rowley1998neural} and boosted decision trees \cite{dollar}. This algorithm usually starts with a dataset of positive examples and a random set of negative examples. The machine learning model is then trained to convergence on that dataset and subsequently applied to a larger dataset to harvest false positives. The false positives are then added to the training set and then the model is trained again. This process is usually iterated only once and does not have any convergence proofs.

\vspace{-0.1in}
\paragraph{ConvNet-based object detection.} 
In the last three years significant gains have been made in object detection. These improvements were made possible by the successful application of deep ConvNets \cite{AlexNet} to ImageNet classification \cite{imagenet}. The R-CNN \cite{rcnn} and OverFeat \cite{overfeat} detectors lead this wave with impressive results on PASCAL VOC \cite{voc} and ImageNet detection. OverFeat is based on the sliding-window detection method, which is perhaps the most intuitive and oldest search method for detection. R-CNN, in contrast, uses region proposals \cite{alexe2010objec,objectness,endres2010category,Uijlings13, cpmc,mcg,BingObj2014,ZitnickEdgeBoxes2014,krahenbuhl2014geodesic}, a method that was made popular by the selective search algorithm \cite{Uijlings13}. Since R-CNN, there has been rapid progress in region-based ConvNets, including SPPnet \cite{SPPnet}, MR-CNN \cite{mrcnn}, and Fast R-CNN \cite{frcn}, which our work builds on.

\vspace{-0.1in}
\paragraph{Hard example selection in deep learning.}
There is recent work \cite{simo2014fracking,wang2015unsupervised,loshchilov2015online} concurrent to our own that selects hard examples for training deep networks. Similar to our approach, all these methods base their selection on the current loss for each datapoint. \cite{simo2014fracking} independently selects hard positive and negative example from a larger set of random examples based on their loss to learn image descriptors. Given a positive pair of patches,~\cite{wang2015unsupervised} finds hard negative patches from a large set using triplet loss. Akin to our approach,~\cite{loshchilov2015online} investigates online selection of hard examples for mini-batch SGD methods. Their selection is also based on loss, but the focus is on ConvNets for image classification. Complementary to~\cite{loshchilov2015online}, we focus on online hard example selection strategy for region-based object detectors.


\section{Overview of Fast R-CNN}\label{sec:frcn-overview-sec}
\vspace{-0.05in}
We first summarize the Fast R-CNN~\cite{frcn} (FRCN) framework. FRCN takes as input an image and a set of object proposal regions of interest (RoIs). The FRCN network itself can be divided into two sequential parts: a convolutional (\emph{conv}) network with several convolution and max-pooling layers (Figure~\ref{fig:frcn-overview}, ``Convolutional Network''); and an RoI network with an RoI-pooling layer, several fully-connected (\emph{fc}) layers and two loss layers (Figure~\ref{fig:frcn-overview}, ``RoI Network'').

During inference, the conv network is applied to the given image to produce a conv feature map, size of which depends on the input image dimensions. Then, for each object proposal, the RoI-pooling layer projects the proposal onto the conv feature map and extracts a fixed-length feature vector. Each feature vector is fed into the fc layers, which finally give two outputs: (1) a softmax probability distribution over the object classes and background; and (2) regressed coordinates for bounding-box relocalization.

There are several reasons for choosing FRCN as our base object detector, apart from it being a fast end-to-end system. Firstly, the basic two network setup (conv and RoI) is also used by other recent detectors like SPPnet and MR-CNN; therefore, our proposed algorithm is more broadly applicable. Secondly, though the basic setup is similar, FRCN also allows for training the entire conv network, as opposed to both SPPnet and MR-CNN which keep the conv network fixed. And finally, both SPPnet and MR-CNN require features from the RoI network to be cached for training a separate SVM classifier (using hard negative mining). FRCN uses the RoI network itself to train the desired classifiers. In fact, \cite{frcn} shows that in the unified system using the SVM classifiers at later stages was unnecessary.



\subsection{Training}\label{sec:mini}
Like most deep networks, FRCN is trained using stochastic gradient descent (SGD). The loss per example RoI is the sum of a classification log loss that encourages predicting the correct object (or background) label and a localization loss that encourages predicting an accurate bounding box (see \cite{frcn} for details).

To share conv network computation between RoIs, SGD mini-batches are created hierarchically. For each mini-batch, $N$ images are first sampled from the dataset, and then $B/N$ RoIs are sampled from each image. Setting $N=2$ and $B=128$ works well in practice \cite{frcn}. The RoI sampling procedure uses several heuristics, which we describe briefly below.
One contribution of this paper is to eliminate some of these heuristics and their hyperparameters.

\vspace{-0.1in}
\paragraph{Foreground RoIs.} For an example RoI to be labeled as foreground (\texttt{fg}), its intersection over union (IoU) overlap with a ground-truth bounding box should be at least $0.5$. This is a fairly standard design choice, in part inspired by the evaluation protocol of the PASCAL VOC object detection benchmark. The same criterion is used in the SVM hard mining procedures of R-CNN, SPPnet, and MR-CNN. We use the same setting.

\vspace{-0.1in}
\paragraph{Background RoIs.}  A region is labeled background (\texttt{bg}) if its maximum IoU with ground truth is in the interval $[$\texttt{bg\_lo}$, 0.5)$. A lower threshold of \texttt{bg\_lo} $=0.1$ is used by both FRCN and SPPnet, and is hypothesized in \cite{frcn} to crudely approximate hard negative mining; the assumption is that regions with some overlap with the ground truth are more likely to be the confusing or hard ones. We show in Section~\ref{sec:allroi} that although this heuristic helps convergence and detection accuracy, it is suboptimal because it ignores some infrequent, but important, difficult background regions. Our method removes the \texttt{bg\_lo} threshold.

\vspace{-0.1in}
\paragraph{Balancing \texttt{fg}-\texttt{bg} RoIs:}
To handle the data imbalance described in Section~\ref{sec:intro},~\cite{frcn} designed heuristics to rebalance the foreground-to-background ratio in each mini-batch to a target of $1:3$ by undersampling the background patches at random, thus ensuring that $25\%$ of a mini-batch is \texttt{fg} RoIs. We found that this is an important design decision for the training FRCN. Removing this ratio (\ie randomly sampling RoIs), or increasing it, decreases accuracy by ${\sim}3$ points mAP. With our proposed method, we can remove this ratio hyperparameter with no ill effect.


\begin{figure*}[t]
	\centering
	\includegraphics[width=\linewidth]{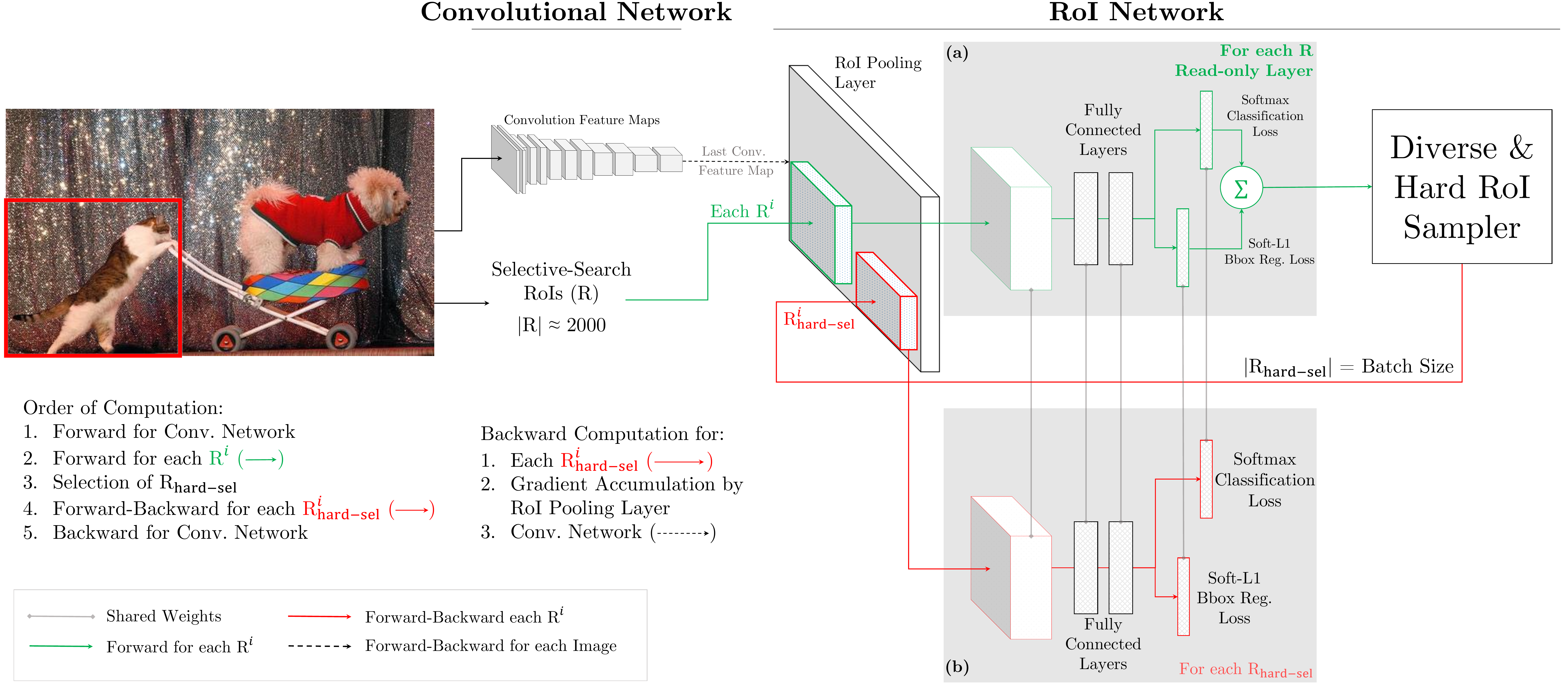}
	\vspace{-0.25in}
  \caption[caption]{Architecture of the proposed training algorithm. Given an image, and selective search RoIs, the conv network computes a conv feature map. In (a), the \protect\emph{readonly} RoI network runs a forward pass on the feature map and all RoIs (shown in {\protect\color{green}green} arrows). Then the Hard RoI module uses these RoI losses to select $B$ examples. In (b), these hard examples are used by the RoI network to compute forward and backward passes (shown in {\protect\color{red}red} arrows).}
\vspace{-0.2in}
	\label{fig:overview}
\end{figure*}

\section{Our approach}
We propose a simple yet effective online hard example mining algorithm for training Fast R-CNN (or any Fast R-CNN style object detector). We argue that the current way of creating mini-batches for SGD (Section \ref{sec:mini}) is inefficient and suboptimal, and we demonstrate that our approach leads to better training (lower training loss) and higher testing performance (mAP).

\subsection{Online hard example mining}\label{sec:ohem}
Recall the alternating steps that define a hard example mining algorithm: (a) for some period of time a \emph{fixed} model is used to find new examples to add to the active training set; (b) then, for some period of time the model is trained on the \emph{fixed} active training set. In the context of SVM-based object detectors, such as the SVMs trained in R-CNN or SPPnet, step (a) inspects a variable number of images (often 10's or 100's) until the active training set reaches a threshold size, and then in step (b) the SVM is trained to convergence on the active training set. This process repeats until the active training set contains all support vectors. Applying an analogous strategy to FRCN ConvNet training slows learning because no model updates are made while selecting examples from the 10's or 100's of images.

Our main observation is that these alternating steps can be combined with how FRCN is trained using online SGD. The key is that although each SGD iteration samples only a small number of images, each image contains \emph{thousands} of example RoIs from which we can select the hard examples rather than a heuristically sampled subset. This strategy fits the alternation template to SGD by ``freezing'' the model for only one mini-batch. Thus the model is updated exactly as frequently as with the baseline SGD approach and therefore learning is not delayed.

More specifically, the online hard example mining algorithm (OHEM) proceeds as follows. For an input image at SGD iteration $t$, we first compute a conv feature map using the conv network. Then the RoI network uses this feature map and the \textbf{all} the input RoIs $(\mathrm{R})$, instead of a sampled mini-batch~\cite{frcn}, to do a forward pass. Recall that this step only involves RoI pooling, a few fc layers, and loss computation for each RoI. The loss represents how well the current network performs on each RoI. Hard examples are selected by sorting the input RoIs by loss and taking the $B/N$ examples for which the current network performs worst. Most of the forward computation is shared between RoIs via the conv feature map, so the extra computation needed to forward all RoIs is relatively small. Moreover, because only a small number of RoIs are selected for updating the model, the backward pass is no more expensive than before.

However, there is a small caveat: co-located RoIs with high overlap are likely to have correlated losses. Moreover, these overlapping RoIs can project onto the same region in the conv feature map, because of resolution disparity, thus leading to loss double counting. To deal with these redundant and correlated regions, we use standard non-maximum suppression (NMS) to perform deduplication (the implementation from \cite{frcn}). Given a list of RoIs and their losses, NMS works by iteratively selecting the RoI with the highest loss, and then removing all lower loss RoIs that have high overlap with the selected region. We use a relaxed IoU threshold of $0.7$ to suppress only highly overlapping RoIs.

We note that the procedure described above does not need a \texttt{fg}-\texttt{bg} ratio for data balancing. If any class were neglected, its loss would increase until it has a high probability of being sampled. There can be images where the \texttt{fg} RoIs are easy (\eg canonical view of a car), so the network is free to use only \texttt{bg} regions in a mini-batch; and vice-versa when \texttt{bg} is trivial (\eg sky, grass \etc), the mini-batch can be entirely \texttt{fg} regions.

\subsection{Implementation details}\label{sec:details}
There are many ways to implement OHEM in the FRCN detector, each with different trade-offs. An obvious way is to modify the loss layers to do the hard example selection. The loss layer can compute loss for all RoIs, sort them based on this loss to select \emph{hard} RoIs, and finally set the loss of all \emph{non-hard} RoIs to $0$. Though straightforward, this implementation is inefficient as the RoI network still allocates memory and performs backward pass for \textbf{all} RoIs, even though most RoIs have $0$ loss and hence no gradient updates (a limitation of current deep learning toolboxes).

To overcome this, we propose the architecture presented in Figure~\ref{fig:overview}. Our implementation maintains two copies of the RoI network, one of which is \emph{readonly}. This implies that the readonly RoI network (Figure~\ref{fig:overview}(a)) allocates memory only for forward pass of all RoIs as opposed to the standard RoI network, which allocates memory for both forward and backward passes. For an SGD iteration, given the conv feature map, the readonly RoI network performs a  forward pass and computes loss for \textbf{all} input RoIs $\left(\mathrm{R}\right)$ (Figure~\ref{fig:overview}, {\color{green} green} arrows). Then the hard RoI sampling module uses the procedure described in Section~\ref{sec:ohem} to select hard examples $\left(\mathrm{R_\text{hard-sel}}\right)$, which are input to the regular RoI network (Figure~\ref{fig:overview}(b), {\color{red} red} arrows)). This network computes forward and backward passes only for $\mathrm{R_\text{hard-sel}}$, accumulates the gradients and passes them to the conv network. In practice, we use all RoIs from all $N$ images as $\mathrm{R}$, therefore the effective batch size for the readonly RoI network is $|\mathrm{R}|$ and for the regular RoI network is the standard $B$ from Section~\ref{sec:mini}.

We implement both options described above using the Caffe~\cite{caffe} framework (see~\cite{frcn}). Our implementation uses gradient accumulation with $N$ forward-backward passes of single image mini-batches. Following FRCN~\cite{frcn}, we use $N=2$ (which results in $|\mathrm{R}| \approx 4000$) and $B=128$. Under these settings, the proposed architecture (Figure~\ref{fig:overview}) has similar memory footprint as the first option, but is $>2\times$ faster. Unless specified otherwise, the architecture and settings described above will be used throughout this paper.



\vspace{-0.05in}
\section{\label{sec:analyze}Analyzing online hard example mining}
\vspace{-0.03in}
This section compares FRCN training with online hard example mining (OHEM) to the baseline heuristic sampling approach. We also compare FRCN with OHEM to a less efficient approach that uses all available example RoIs in each mini-batch, not just the $B$ hardest examples.

\vspace{-0.01in}
\subsection{Experimental setup}\label{sec:expsetup}
\vspace{-0.03in}
We conduct experiments with two standard ConvNet architectures: VGG\_CNN\_M\_1024 (VGGM, for short) from~\cite{Chatfield14}, which is a wider version of AlexNet~\cite{AlexNet}, and VGG16 from~\cite{VGG}. All experiments in this section are performed on the PASCAL VOC07 dataset. Training is done on the trainval set and testing on the test set. Unless specified otherwise, we will use the default settings from FRCN~\cite{frcn}. We train all methods with SGD for 80k mini-batch iterations, with an initial learning rate of 0.001 and we decay the learning rate by 0.1 every 30k iterations. The baseline numbers reported in Table~\ref{tab:ablation} (row 1-2) were reproduced using our training schedule and are slightly higher than the ones reported in~\cite{frcn}.

\subsection{OHEM vs.\ heuristic sampling}\label{sec:bglo}
\vspace{-0.03in}
Standard FRCN, reported in Table~\ref{tab:ablation} (rows $1-2$), uses $\texttt{bg\_lo}=0.1$ as a heuristic for hard mining (Section~\ref{sec:mini}). To test the importance of this heuristic, we ran FRCN with $\texttt{bg\_lo}=0$. Table~\ref{tab:ablation} (rows $3-4$) shows that for VGGM, mAP drops by $2.4$ points, whereas for VGG16 it remains roughly the same. Now compare this to training FRCN with OHEM (rows $11-13$). OHEM improves mAP by $2.4$ points compared to FRCN with the $\texttt{bg\_lo}=0.1$ heuristic for VGGM, and $4.8$ points without the heuristic. This result demonstrates the sub-optimality of these heuristics and the effectiveness of our hard mining approach.

\subsection{Robust gradient estimates}\label{sec:fewer}
\vspace{-0.03in}
One concern over using only $N=2$ images per batch is that it may cause unstable gradients and slow convergence because RoIs from an image may be highly correlated~\cite{minibatchSVM}. FRCN~\cite{frcn} reports that this was not a practical issue for their training. But this detail might raise concerns over our training procedure because we use examples with high loss from the same image and as a result they may be more highly correlated. To address this concern, we experiment with $N=1$ in order to increase correlation in an effort to break our method. As seen in Table~\ref{tab:ablation} (rows $5-6,11$), performance of the original FRCN drops by ${\sim}1$ point with $N=1$, but when using our training procedure, mAP remains approximately the same. This shows that OHEM is robust in case one needs fewer images per batch in order to reduce GPU memory usage.

\subsection{Why just hard examples, when you can use all?}\label{sec:allroi}
Online hard example mining is based on the hypothesis that it is important to consider all RoIs in an image and then select hard examples for training. But what if we train with all the RoIs, not just the hard ones? The easy examples will have low loss, and won't contribute much to the gradient; training will automatically focus on the hard examples. To compare this option, we ran standard FRCN training with a large mini-batch size of $B = 2048$, using $\texttt{bg{\_}lo}=0$, $N\in\{1,2\}$ and with other hyperparameters fixed. Because this experiment uses a large mini-batch, it's important to tune the learning rate to adjust for this change. We found optimal results by increasing it to $0.003$ for VGG16 and $0.004$ for VGGM. The outcomes are reported in Table~\ref{tab:ablation} (rows $7-10$). Using these settings, mAP of both VGG16 and VGGM increased by ${\sim}1$ point compared to $B=128$, but the improvement from our approach is still $>1$ points over using all RoIs. Moreover, because we compute gradients with a smaller mini-batch size training is faster.

\newcolumntype{G}{>{\color{gray!100}}c<{}}
\newcolumntype{L}[1]{>{\raggedleft\let\newline\\\arraybackslash\hspace{0pt}\color{gray!100}}m{#1}<{}}
\begin{table}[t]
\centering
\scriptsize
\setlength{\tabcolsep}{0.6em}
\caption[caption]{Impact of hyperparameters on FRCN training.}
\vspace{-0.2in}
\center
\begin{tabular}{@{}L{0.3cm} c c c c c c c c c c}
\toprule
& Experiment & Model & $N$ & LR & $B$ & \texttt{bg\_lo} & 07 mAP\\
\midrule
1 & \multirow{2}{*}{Fast R-CNN~\cite{frcn}} & VGGM & \multirow{2}{*}{2} & \multirow{2}{*}{0.001} & \multirow{2}{*}{128} & \multirow{2}{*}{0.1} & 59.6\\
2 & & VGG16 & & & & & 67.2\\
\midrule
3 & \multirow{2}{*}{\parbox{2.5cm}{\centering Removing hard mining heuristic (Section~\ref{sec:bglo})}} 
& VGGM & \multirow{2}{*}{2} & \multirow{2}{*}{0.001}
& \multirow{2}{*}{128} & \multirow{2}{*}{\textbf{0}} & 57.2\\
4 & & VGG16 & & & & & 67.5\\
\midrule
5 & \multirow{2}{*}{\parbox{2.5cm}{\centering Fewer images per batch (Section~\ref{sec:fewer})}} 
& \multirow{2}{*}{VGG16} & \multirow{2}{*}{\textbf{1}} & \multirow{2}{*}{0.001}
& \multirow{2}{*}{128} & 0.1 & 66.3\\
6 & & & & & & 0 & 66.3\\
\midrule
7 & \multirow{4}{*}{\parbox{2.5cm}{\centering Bigger batch, High LR (Section~\ref{sec:allroi})}} & \multirow{2}{*}{VGGM} & 1 & \multirow{2}{*}{\textbf{0.004}} & \multirow{2}{*}{\textbf{2048}} & \multirow{2}{*}{0} & 57.7\\
8 & & & 2 & & & & 60.4\\
\arrayrulecolor{Gray}
\cmidrule{3-8}
\arrayrulecolor{black}
9 & & \multirow{2}{*}{VGG16} & 1 & \multirow{2}{*}{\textbf{0.003}} & \multirow{2}{*}{\textbf{2048}} & \multirow{2}{*}{0} & 67.5\\
10 & & & 2 & & & & 68.7\\
\midrule
11 & \multirow{3}{*}{\parbox{1.6cm}{\centering Our Approach}}  & VGG16 & \textbf{1} & 0.001 & 128 & 0 & 69.7\\
\arrayrulecolor{Gray}
\cmidrule{3-8}
\arrayrulecolor{black}
12 & 
 & VGGM & \multirow{2}{*}{2} 
& \multirow{2}{*}{0.001} & \multirow{2}{*}{128} & \multirow{2}{*}{0} & 62.0\\
13 & & VGG16 & & & & & 69.9\\
\bottomrule
\end{tabular}
\label{tab:ablation}
\vspace{-0.15in}
\end{table}

\subsection{Better optimization}
Finally, we analyze the training loss for the various FRCN training methods discussed above. It's important to measure training loss in a way that does not depend on the sampling procedure and thus results in a valid comparison between methods. To achieve this goal, we take model snapshots from each method every 20k steps of optimization and run them over the entire VOC07 trainval set to compute the average loss over \emph{all} RoIs. This measures the training set loss in a way that does not depend on the example sampling scheme.

\begin{figure}[t]
	\centering
	\includegraphics[width=0.235\textwidth]{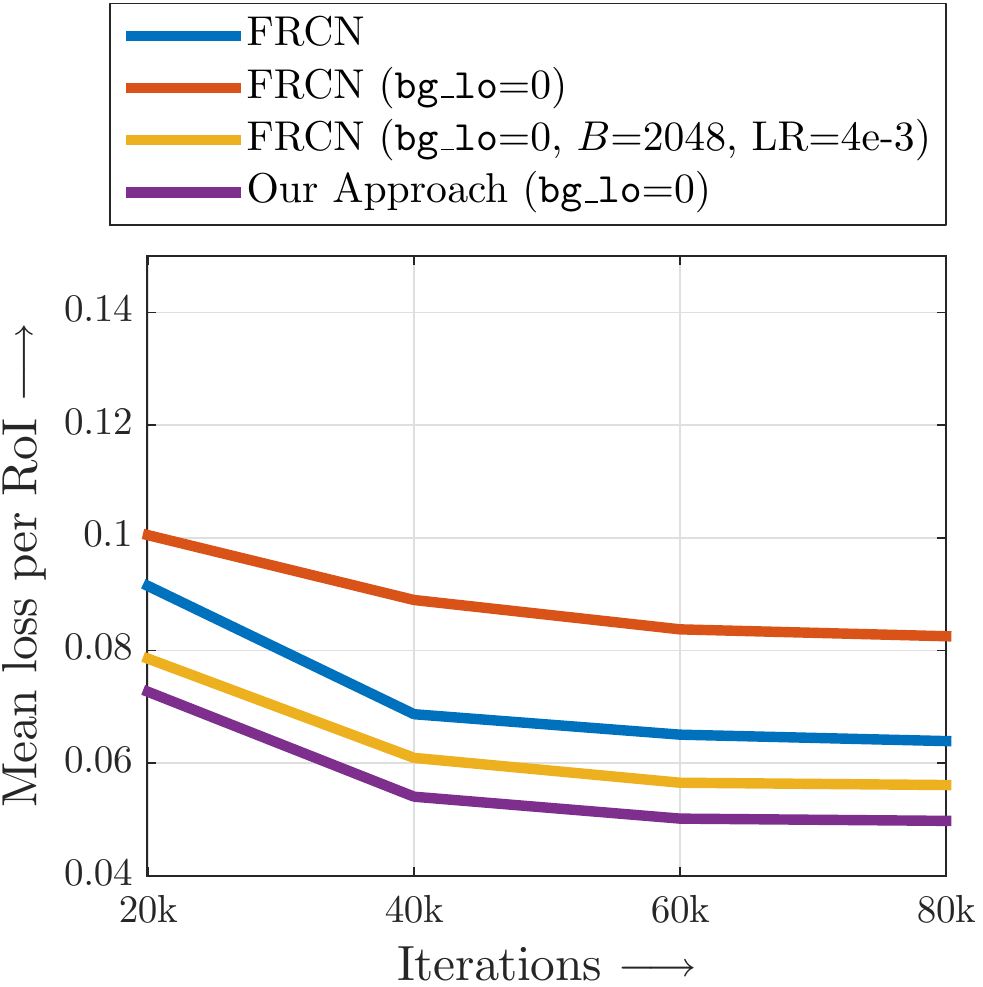}
	\includegraphics[width=0.235\textwidth]{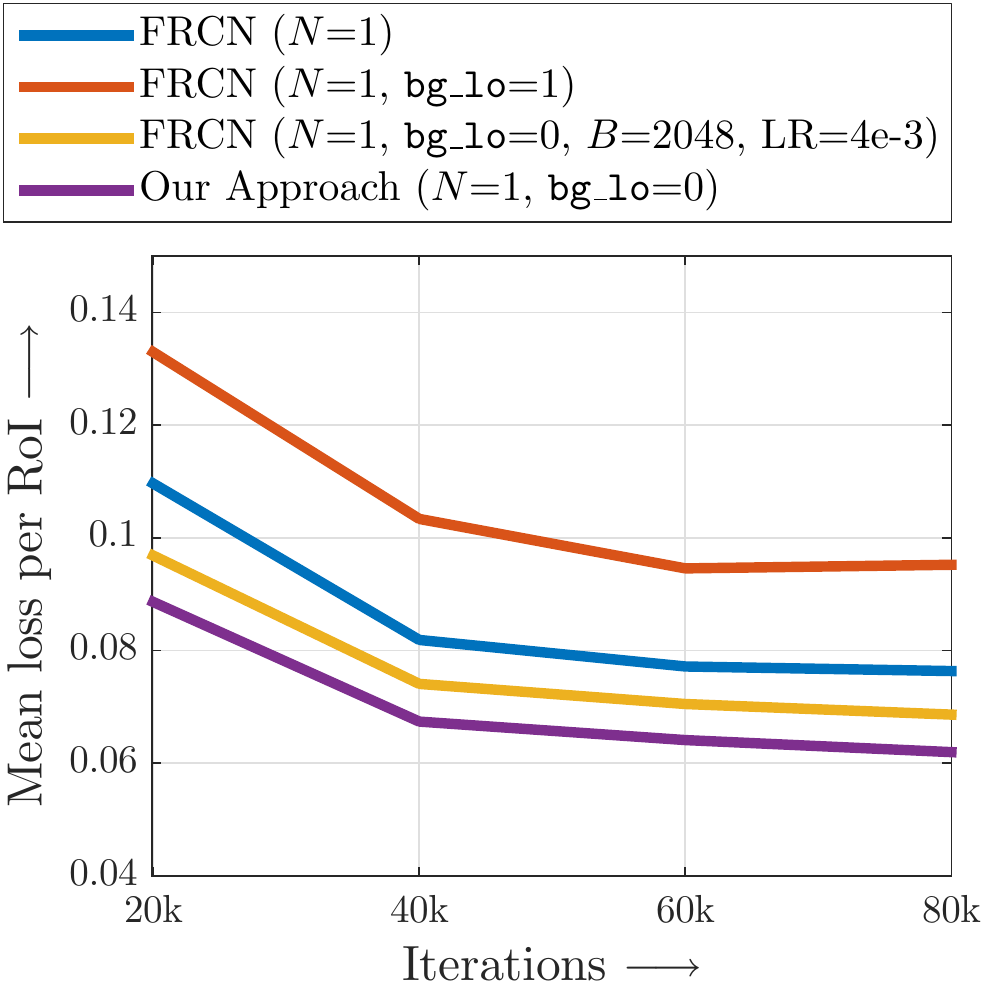}
	\caption[caption]{Training loss is computed for various training procedures using VGG16 networks discussed in Section \protect\ref{sec:analyze}. We report mean loss per RoI. These results indicate that using hard mining for training leads to lower training loss than any of the other heuristics.}
	\label{fig:lossgraph}
\end{figure}

Figure~\ref{fig:lossgraph} shows the average loss per RoI for VGG16 with the various hyperparameter settings discussed above and presented in Table~\ref{tab:ablation}. We see that $\texttt{bg\_lo}=0$ results in the highest training loss, while using the heuristic $\texttt{bg\_lo}=0.1$ results in a much lower training loss. Increasing the mini-batch size to $B=2048$ and increasing the learning rate lowers the training loss below the $\texttt{bg\_lo}=0.1$ heuristic. Our proposed online hard example mining method achieves the lowest training loss of all methods, validating our claims that OHEM leads to better training for FRCN.

\begin{table}[t]
	\centering
	\small
	\caption[caption]{Computational statistics of training FRCN \protect\cite{frcn} and FRCN with OHEM (using an Nvidia Titan X GPU).}
	\vspace{-0.05in}
	\begin{tabular}{@{}l c c c c c@{}}
		\toprule
		& \multicolumn{2}{c}{VGGM} & \multicolumn{3}{c}{VGG16}\\
		\cmidrule(l{1pt}r{2pt}){2-3}\cmidrule(l{2pt}){4-6}
		& FRCN & Ours & FRCN & FRCN* & Ours*\\
		\midrule
		time (sec/iter)& 0.13 & 0.22 & 0.60 & 0.57 & 1.00 \\
		max. memory (G) & 2.6 & 3.6 & 11.2 & 6.4 & 8.7 \\
		\bottomrule
	\end{tabular}
	{\small *: uses gradient accumulation over two forward/backward passes}
	\vspace{-0.1in}
	\label{tab:stats}
\end{table}

\subsection{Computational cost}\label{sec:compstats}
OHEM adds reasonable computational and memory overhead, as reported in Table \ref{tab:stats}. OHEM costs 0.09s per training iteration for VGGM network (0.43s for VGG16) and requires 1G more memory (2.3G for VGG16). Given that FRCN~\cite{frcn} is a fast detector to train, the increase in training time is likely acceptable to most users.



\newcolumntype{x}{>\small c}
\newcolumntype{L}[1]{>{\raggedright\let\newline\\\arraybackslash\hspace{0pt}}m{#1}}
\newcolumntype{C}[1]{>{\centering\let\newline\\\arraybackslash\hspace{0pt}}m{#1}}
\newcolumntype{R}[1]{>{\raggedleft\let\newline\\\arraybackslash\hspace{0pt}}m{#1}}
\begin{table*}[t]
\centering
\caption[caption]{\small {\bf VOC 2007 test} detection average precision (\%). All methods use VGG16. Training set key: {\bf 07}: VOC07 trainval, 
{\bf 07$+$12}: union of {\bf 07} and VOC12 trainval. All methods use bounding-box regression. Legend: \textbf{M}: using multi-scale for training and testing, \textbf{B}: multi-stage bbox regression. FRCN\raisebox{0.2ex}{$\star$} refers to FRCN \protect\cite{frcn} with our training schedule.}
\vspace{-0.1in}
\renewcommand{\arraystretch}{1.2}
\renewcommand{\tabcolsep}{1.2mm}
\resizebox{\linewidth}{!}{
\begin{tabular}{@{}L{2.1cm} C{0.4cm} C{0.3cm} !{\color{gray}\vrule} L{1.2cm} !{\color{gray}\vrule} l !{\color{gray}\vrule} r*{19}{x} @{}}
\Xhline{1pt}
method & \textbf{M} & \textbf{B} & train set & mAP & aero      & bike      & bird      & boat      & bottle     & bus        & car        & cat        & chair      & cow        & table      & dog        & horse      & mbike      & persn     & plant      & sheep      & sofa       & train      & tv   \\
\Xhline{0.8pt}
FRCN~\cite{frcn} & & & 07 & 66.9 & 74.5 & 78.3 & 69.2 & 53.2 & 36.6 & 77.3 & 78.2 & 82.0 & 40.7 & 72.7 & 67.9 & 79.6 & 79.2 & 73.0 & 69.0 & 30.1 & 65.4 & 70.2 & 75.8 & 65.8 \\
FRCN\raisebox{0.2ex}{$\star$} & & & 07 & 67.2 & 74.6 & 76.8 & 67.6 & 52.9 & 37.8 & 78.7 & 78.8 & 81.6 & 42.2 & 73.6 & 67.0 & 79.4 & 79.6 & 74.1 & 68.3 & 33.4 & 65.9 & 68.7 & 75.4 & 68.1 \\
\textbf{Ours} & & & 07 & \textbf{69.9} & 71.2 & 78.3 & 69.2 & 57.9 & 46.5 & 81.8 & 79.1 & 83.2 & 47.9 & 76.2 & 68.9 & 83.2 & 80.8 & 75.8 & 72.7 & 39.9 & 67.5 & 66.2 & 75.6 & 75.9 \\
\Xhline{0.5pt}
FRCN\raisebox{0.2ex}{$\star$} & \checkmark & \checkmark & 07 & 72.4 & 77.8 & 81.3 & 71.4 & 60.4 & 48.3 & 85.0 & 84.6 & 86.2 & 49.4 & 80.7 & 68.1 & 84.1 & 86.7 & 80.2 & 75.3 & 38.7 & 71.9 & 71.5 & 77.9 & 67.8 \\ 
MR-CNN~\cite{mrcnn} & \checkmark & \checkmark & 07 & 74.9 & 78.7 & 81.8 & 76.7 & 66.6 & 61.8 & 81.7 & 85.3 & 82.7 & 57.0 & 81.9 & 73.2 & 84.6 & 86.0 & 80.5 & 74.9 & 44.9 & 71.7 & 69.7 & 78.7 & 79.9 \\ 
\textbf{Ours} & \checkmark & \checkmark & 07 & \textbf{75.1} & 77.7 & 81.9 & 76.0 & 64.9 & 55.8 & 86.3 & 86.0 & 86.8 & 53.2 & 82.9 & 70.3 & 85.0 & 86.3 & 78.7 & 78.0 & 46.8 & 76.1 & 72.7 & 80.9 & 75.5 \\
\Xhline{0.5pt}
FRCN~\cite{frcn} & & & 07$+$12 & 70.0 & 77.0 & 78.1 & 69.3 & 59.4 & 38.3 & 81.6 & 78.6 & 86.7 & 42.8 & 78.8 & 68.9 & 84.7 & 82.0 & 76.6 & 69.9 & 31.8 & 70.1 & 74.8 & 80.4 & 70.4 \\ 
\textbf{Ours} & & & 07$+$12 & \textbf{74.6} & 77.7 & 81.2 & 74.1 & 64.2 & 50.2 & 86.2 & 83.8 & 88.1 & 55.2 & 80.9 & 73.8 & 85.1 & 82.6 & 77.8 & 74.9 & 43.7 & 76.1 & 74.2 & 82.3 & 79.6 \\
\Xhline{0.5pt}
MR-CNN~\cite{mrcnn} & \checkmark & \checkmark & 07$+$12 & 78.2 & 80.3 & 84.1 & 78.5 & \hl{70.8} & \hl{68.5} & 88.0 & 85.9 & 87.8 & \hl{60.3} & \hl{85.2} & 73.7 & \hl{87.2} & 86.5 & \hl{85.0} & 76.4 & 48.5 & 76.3 & 75.5 & \hl{85.0} & \hl{81.0} \\ 
\textbf{Ours} & \checkmark & \checkmark & 07$+$12 & \hl{78.9} & \hl{80.6} & \hl{85.7} & \hl{79.8} & 69.9 & 60.8 & \hl{88.3} & \hl{87.9} & \hl{89.6} & 59.7 & 85.1 & \hl{76.5} & 87.1 & \hl{87.3} & 82.4 & \hl{78.8} & \hl{53.7} & \hl{80.5} & \hl{78.7} & 84.5 & 80.7 \\ 
\Xhline{1pt}
\end{tabular}
}
\vspace{-0.05in}
\label{tab:voc2007}
\end{table*}

\begin{table*}[t]
\centering
\renewcommand{\arraystretch}{1.2}
\renewcommand{\tabcolsep}{1.2mm}
\caption[caption]{{\small {\bf VOC 2012 test} detection average precision (\%). All methods use VGG16. Training set key: {\bf 12}: VOC12 trainval, {\bf 07$++$12}: union of VOC07 trainval, VOC07 test, and VOC12 trainval. Legend: \textbf{M}: using multi-scale for training and testing, \textbf{B}: iterative bbox regression.}}
\vspace{-0.1in}
\resizebox{\linewidth}{!}{
\begin{tabular}{@{}L{2.1cm} C{0.4cm} C{0.3cm} !{\color{gray}\vrule} L{1.3cm} !{\color{gray}\vrule} l !{\color{gray}\vrule} r*{19}{x} @{}}
\Xhline{1pt}
method & \textbf{M} & \textbf{B} & train set & mAP & aero      & bike      & bird      & boat      & bottle     & bus        & car        & cat        & chair      & cow        & table      & dog        & horse      & mbike      & persn     & plant      & sheep      & sofa       & train      & tv  \\
\Xhline{1pt}
FRCN~\cite{frcn} & & & 12 & 65.7 & 80.3 & 74.7 & 66.9 & 46.9 & 37.7 & 73.9 & 68.6 & 87.7 & 41.7 & 71.1 & 51.1 & 86.0 & 77.8 & 79.8 & 69.8 & 32.1 & 65.5 & 63.8 & 76.4 & 61.7 \\
{\bf Ours$^1$} & & & 12  & \textbf{69.8} & 81.5 & 78.9 & 69.6 & 52.3 & 46.5 & 77.4 & 72.1 & 88.2 & 48.8 & 73.8 & 58.3 & 86.9 & 79.7 & 81.4 & 75.0 & 43.0 & 69.5 & 64.8 & 78.5 & 68.9 \\
\Xhline{0.5pt}
MR-CNN~\cite{mrcnn} & \checkmark & \checkmark & 12  & 70.7 & 85.0 & 79.6 & 71.5 & 55.3 & 57.7 & 76.0 & 73.9 & 84.6 & 50.5 & 74.3 & 61.7 & 85.5 & 79.9 & 81.7 & 76.4 & 41.0 & 69.0 & 61.2 & 77.7 & 72.1 \\
{\bf Ours$^2$} &\checkmark & \checkmark & 12 & \textbf{72.9} & 85.8 & 82.3 & 74.1 & 55.8 & 55.1 & 79.5 & 77.7 & 90.4 & 52.1 & 75.5 & 58.4 & 88.6 & 82.4 & 83.1 & 78.3 & 47.0 & 77.2 & 65.1 & 79.3 & 70.4\\
\Xhline{0.5pt}
FRCN~\cite{frcn} & & & 07$++$12  & 68.4 & 82.3 & 78.4 & 70.8 & 52.3 & 38.7 & 77.8 & 71.6 & 89.3 & 44.2 & 73.0 & 55.0 & 87.5 & 80.5 & 80.8 & 72.0 & 35.1 & 68.3 & 65.7 & 80.4 & 64.2
\\			
{\bf Ours}$^3$ & & & 07$++$12  & \textbf{71.9} & 83.0 & 81.3 & 72.5 & 55.6 & 49.0 & 78.9 & 74.7 & 89.5 & 52.3 & 75.0 & 61.0 & 87.9 & 80.9 & 82.4 & 76.3 & 47.1 & 72.5 & 67.3 & 80.6 & 71.2 \\
\Xhline{0.5pt}
MR-CNN~\cite{mrcnn} & \checkmark & \checkmark & 07$++$12 & 73.9 &  85.5 & 82.9 & 76.6 & 57.8 & \hl{62.7} & 79.4 & 77.2 & 86.6 & 55.0 & 79.1 & 62.2 & 87.0 & 83.4 & 84.7 & 78.9 & 45.3 & 73.4 & 65.8 & 80.3 & 74.0 \\
{\bf Ours}$^4$ & \checkmark & \checkmark & 07$++$12  & \hl{76.3} & \hl{86.3} & \hl{85.0} & \hl{77.0} & \hl{60.9} & 59.3 & \hl{81.9} & \hl{81.1} & \hl{91.9} & \hl{55.8} & \hl{80.6} & \hl{63.0} & \hl{90.8} & \hl{85.1} & \hl{85.3} & \hl{80.7} & \hl{54.9} & \hl{78.3} & \hl{70.8} & \hl{82.8} & \hl{74.9} \\
\Xhline{1pt}
\end{tabular}
}
{\scriptsize $^1$\url{http://host.robots.ox.ac.uk:8080/anonymous/XNDVK7.html}, $^2$\url{http://host.robots.ox.ac.uk:8080/anonymous/H49PTT.html}, $^3$\url{http://host.robots.ox.ac.uk:8080/anonymous/LSANTB.html}, $^4$\url{http://host.robots.ox.ac.uk:8080/anonymous/R7EAMX.html}}
	\vspace{-0.1in}
\label{tab:voc2012}
\end{table*}

\section{PASCAL VOC and MS COCO results}
In this section, we evaluate our method on VOC 2012 \cite{voc} as well as the more challenging MS COCO \cite{coco} dataset.
We demonstrate consistent and significant improvement in FRCN performance when using the proposed OHEM approach.
Per-class results are also presented on VOC 2007 for comparison with prior work.

\vspace{-0.05in}
\paragraph{Experimental setup.} We use VGG16 for all experiments. When training on VOC07 trainval, we use the SGD parameters as in Section~\ref{sec:analyze} and when using extra data (07+12 and 07++12, see Table~\ref{tab:voc2007} and~\ref{tab:voc2012}), we use 200k mini-batch iterations, with an initial learning rate of 0.001 and decay step size of 40k. When training on MS COCO~\cite{coco}, we use 240k mini-batch iterations, with an initial learning rate of 0.001 and decay step size of 160k, owing to a larger epoch size.

\subsection{VOC 2007 and 2012 results}
Table \ref{tab:voc2007} shows that on VOC07, OHEM improves the mAP of FRCN from 67.2\% to 69.9\% (and 70.0\% to 74.6\% with extra data). 
On VOC12, OHEM leads to an improvement of 4.1 points in mAP (from 65.7\% to 69.8\%). With extra data, we achieve an mAP of 71.9\% as compared to 68.4\% mAP of FRCN, an improvement of 3.5 points. Interestingly the improvements are not uniform across categories. Bottle, chair, and tvmonitor show larger improvements that are consistent across the different PASCAL splits. Why these classes benefit the most is an interesting and open question.

\subsection{MS COCO results}
To test the benefit of using OHEM on a larger and more challenging dataset, we conduct experiments on MS COCO~\cite{coco} and report numbers from test-dev 2015 evaluation server (Table~\ref{tab:coco}). On the standard COCO evaluation metric, FRCN~\cite{frcn} scores 19.7\% AP, and OHEM improves it to 22.6\% AP.\footnote{COCO AP averages over classes, recall, and IoU levels. See \url{http://mscoco.org/dataset/\#detections-eval} for details.} Using the VOC overlap metric of $\text{IoU}\ge0.5$, OHEM gives a 6.6 points boost in AP$^{50}$. It is also interesting to note that OHEM helps improve the AP of medium sized objects by 4.9 points on the strict COCO AP evaluation metric, which indicates that the proposed hard example mining approach is helpful when dealing with smaller sized objects. Note that FRCN with and without OHEM were trained on MS COCO train set.

\section{Adding bells and whistles}
We've demonstrated consistent gains in detection accuracy by applying OHEM to FRCN training. In this section, we show that these improvements are orthogonal to recent bells and whistles that enhance object detection accuracy. OHEM with the following two additions yields state-of-the-art results on VOC and competitive results on MS COCO.

\vspace{-0.15in}
\paragraph{Multi-scale (M).} We adopt the multi-scale strategy from SPPnet~\cite{SPPnet} (and used by both FRCN~\cite{frcn} and MR-CNN~\cite{mrcnn}). Scale is defined as the size of the shortest side ($s$) of an image. During training, one scale is chosen at random, whereas at test time inference is run on all scales. For VGG16 networks, we use $s \in \{480, 576, 688, 864, 900\}$ for training, and $s \in \{480, 576, 688, 864, 1000\}$ during testing, with the max dimension capped at 1000. The scales and caps were chosen because of GPU memory constraints.

\paragraph{Iterative bounding-box regression (B).} We adopt the iterative localization and bounding-box (bbox) voting scheme from~\cite{mrcnn}.
The network evaluates each proposal RoI to get scores and relocalized boxes $\mathrm{R}_1$. High-scoring $\mathrm{R}_1$ boxes are the rescored and relocalized, yielding boxes $\mathrm{R}_2$. Union of $\mathrm{R}_1$ and $\mathrm{R}_2$ is used as the final set $\mathrm{R}_\text{F}$ for post-processing, where $\mathrm{R}^\text{NMS}_\text{F}$ is obtained using NMS on $\mathrm{R}_\text{F}$ with an IoU threshold of 0.3 and weighted voting is performed on each box $\mathrm{r}_i$ in $\mathrm{R}^\text{NMS}_\text{F}$ using boxes in $\mathrm{R}_\text{F}$ with an IoU of $\ge$0.5 with $\mathrm{r}_i$ (see~\cite{mrcnn} for details).

\newcolumntype{L}[1]{>{\raggedright\let\newline\\\arraybackslash\hspace{0pt}}m{#1}}
\newcolumntype{C}[1]{>{\centering\let\newline\\\arraybackslash\hspace{0pt}}m{#1}}
\newcolumntype{R}[1]{>{\raggedleft\let\newline\\\arraybackslash\hspace{0pt}}m{#1}}
\newcolumntype{x}{>\small c}
\begin{table}[t]
\centering
\caption[caption]{\textbf{MS COCO 2015 test$-$dev} detection average precision (\%). All methods use VGG16. Legend: \textbf{M}: using multi-scale for training and testing.}
\vspace{-0.1in}
\renewcommand{\arraystretch}{1.2}
\renewcommand{\tabcolsep}{1.2mm}
\resizebox{\linewidth}{!}{
\begin{tabular}{@{}R{2cm} R{1cm} c c c c@{}}
\toprule
AP$@$IoU & area & FRCN${^\dag}$ & \textbf{Ours} & \textbf{Ours} [+\textbf{M}] & \textbf{Ours*} [+\textbf{M}]\\
\midrule
$[0.50:0.95]$ &    all & 19.7 & 22.6 & 24.4 & 25.5 \\
\arrayrulecolor{Gray}
\midrule
\arrayrulecolor{black}
  $0.50$      &    all & 35.9 & 42.5 & 44.4 & 45.9 \\
  $0.75$      &    all & 19.9 & 22.2 & 24.8 & 26.1\\
$[0.50:0.95]$ &  small & 3.5 & 5.0 & 7.1 & 7.4\\
$[0.50:0.95]$ &   med. & 18.8 & 23.7 & 26.4 & 27.7\\
$[0.50:0.95]$ &  large & 34.6 & 37.9 & 38.5 & 40.3\\
\bottomrule
\end{tabular}
}
\\
${^\dag}$from the leaderboard, *trained on trainval set
\vspace{-0.15in}
\label{tab:coco}
\end{table}

\subsection{VOC 2007 and 2012 results}
We report the results on VOC benchmarks in Table~\ref{tab:voc2007} and~\ref{tab:voc2012}. On VOC07, FRCN with the above mentioned additions achieves 72.4\% mAP and OHEM improves it to \textbf{75.1\%}, which is currently the highest reported score under this setting (07 data). When using extra data (07+12), OHEM achieves \textbf{78.9\%} mAP, surpassing the current state-of-the-art MR-CNN (78.2\% mAP). 
We note that MR-CNN uses selective search and edge boxes during training, whereas we only use selective search boxes. Our multi-scale implementation is also different, using fewer scales than MR-CNN. On VOC12 (Table~\ref{tab:voc2012}), we consistently perform better than MR-CNN. When using extra data, we achieve state-of-the-art mAP of {\bf 76.3\%} (\vs 73.9\% mAP of MR-CNN).

\vspace{-0.07in}
\paragraph{Ablation analysis.} We now study in detail the impact of these two additions and whether OHEM is complementary to them, and report the analysis in Table~\ref{tab:msms}. Baseline FRCN mAP improves from 67.2\% to 68.6\% when using multi-scale during both training and testing (we refer to this as~\textbf{M}). However, note that there is only a marginal benefit of using it at training time. Iterative bbox regression (\textbf{B}) further improves the FRCN mAP to 72.4\%. But more importantly, using OHEM improves it to \textbf{75.1\%} mAP, which is state-of-the-art for methods trained on VOC07 data (see Table~\ref{tab:voc2007}). In fact, using OHEM consistently results in higher mAP for all variants of these two additions (see Table~\ref{tab:msms}).

\begin{table}[t!]
\centering
\small
\setlength{\tabcolsep}{0.8em}
\caption[caption]{Impact of multi-scale and iterative bbox reg.}
\vspace{-0.1in}
\begin{tabular}{cc c c c}
\toprule
\multicolumn{2}{c}{Multi-scale (\textbf{M})} & \multirow{2}{*}{\parbox{2cm}{\centering Iterative bbox reg. (\textbf{B})}} & \multicolumn{2}{c}{VOC07 mAP} \\\cmidrule{1-2}\cmidrule{4-5}
Train & Test &  & FRCN & \textbf{Ours} \\
\midrule
& & & 67.2 & 69.9\\
& $\checkmark$ & & 68.4& 71.1\\
& & $\checkmark$ & 70.8& 72.7\\
& $\checkmark$& $\checkmark$ & 71.9 & 74.1\\
\midrule
$\checkmark$& & & 67.7& 70.7\\
$\checkmark$& $\checkmark$ & & 68.6& 71.9\\
$\checkmark$& & $\checkmark$ & 71.2& 72.9\\
$\checkmark$& $\checkmark$& $\checkmark$ & 72.4 & \textbf{75.1}\\
\bottomrule
\end{tabular}
\label{tab:msms}
\vspace{-0.13in}
\end{table}

\subsection{MS COCO results}
MS COCO~\cite{coco} test-dev 2015 evaluation server results are reported in Table~\ref{tab:coco}. Using multi-scale improves the performance of our method to 24.4\% AP on the standard COCO metric and to 44.4\% AP$^{50}$ on the VOC metric. This again shows the complementary nature of using multi-scale and OHEM. Finally, we train our method using the entire MS COCO trainval set, which further improves performance to \textbf{25.5\%} AP (and 45.9\% AP$^{50}$). In the 2015 MS COCO Detection Challenge, a variant of this approach finished $4^{\text{th}}$ place overall.


\vspace{-0.05in}
\section{Conclusion}
\vspace{-0.03in}
We presented an online hard example mining (OHEM) algorithm, a simple and effective method to train region-based ConvNet detectors. OHEM eliminates several heuristics and hyperparameters in common use by automatically selecting hard examples, thus simplifying training. We conducted extensive experimental analysis to demonstrate the effectiveness of the proposed algorithm, which leads to better training convergence and consistent improvements in detection accuracy on standard benchmarks. We also reported state-of-the-art results on PASCAL VOC 2007 and 2012 when using OHEM with other orthogonal additions. Though we used Fast R-CNN throughout this paper, OHEM can be used for training any region-based ConvNet detector.

Our experimental analysis was based on the overall detection accuracy, however it will be an interesting future direction to study the impact of various training methodologies on individual category performance.

\vspace{-0.1in}
\paragraph{Acknowledgment.}
This project started as an intern project at Microsoft Research and continued at CMU. We thank Larry Zitnick, Ishan Misra and Sean Bell for many helpful discussions. AS was supported by the Microsoft Research PhD Fellowship. This work was also partially supported by ONR MURI N000141612007. We thank NVIDIA for donating GPUs.

\bibliographystyle{ieee}
\bibliography{ref}

\begin{thebibliography}{10}\itemsep=-1pt

\bibitem{alexe2010objec}
B.~Alexe, T.~Deselaers, and V.~Ferrari.
\newblock What is an object?
\newblock In {\em CVPR}, 2010.

\bibitem{objectness}
B.~Alexe, T.~Deselaers, and V.~Ferrari.
\newblock Measuring the objectness of image windows.
\newblock {\em TPAMI}, 2012.

\bibitem{mcg}
P.~Arbel{\'a}ez, J.~Pont-Tuset, J.~T. Barron, F.~Marques, and J.~Malik.
\newblock Multiscale combinatorial grouping.
\newblock In {\em CVPR}, 2014.

\bibitem{cpmc}
J.~Carreira and C.~Sminchisescu.
\newblock Constrained parametric min-cuts for automatic object segmentation.
\newblock In {\em CVPR}, 2010.

\bibitem{Chatfield14}
K.~Chatfield, K.~Simonyan, A.~Vedaldi, and A.~Zisserman.
\newblock Return of the devil in the details: Delving deep into convolutional
  nets.
\newblock In {\em {BMVC}}, 2014.

\bibitem{BingObj2014}
M.-M. Cheng, Z.~Zhang, W.-Y. Lin, and P.~H.~S. Torr.
\newblock {BING}: Binarized normed gradients for objectness estimation at
  300fps.
\newblock In {\em CVPR}, 2014.

\bibitem{DalalTriggs}
N.~Dalal and B.~Triggs.
\newblock Histograms of oriented gradients for human detection.
\newblock In {\em CVPR}, 2005.

\bibitem{imagenet}
J.~Deng, W.~Dong, R.~Socher, L.-J. Li, K.~Li, and L.~Fei-Fei.
\newblock Imagenet: A large-scale hierarchical image database.
\newblock In {\em CVPR}, 2009.

\bibitem{dollar}
P.~Doll\'ar, Z.~Tu, P.~Perona, and S.~Belongie.
\newblock Integral channel features.
\newblock In {\em BMVC}, 2009.

\bibitem{endres2010category}
I.~Endres and D.~Hoiem.
\newblock Category independent object proposals.
\newblock In {\em ECCV}, 2010.

\bibitem{voc}
M.~Everingham, L.~Van~Gool, C.~K.~I. Williams, J.~Winn, and A.~Zisserman.
\newblock The pascal visual object classes (voc) challenge.
\newblock {\em IJCV}, 2010.

\bibitem{dpm}
P.~Felzenszwalb, R.~Girshick, D.~McAllester, and D.~Ramanan.
\newblock Object detection with discriminatively trained part-based models.
\newblock {\em PAMI}, 2010.

\bibitem{mrcnn}
S.~Gidaris and N.~Komodakis.
\newblock Object detection via a multi-region \& semantic segmentation-aware
  cnn model.
\newblock In {\em ICCV}, 2015.

\bibitem{frcn}
R.~Girshick.
\newblock Fast {R}-{CNN}.
\newblock In {\em ICCV}, 2015.

\bibitem{rcnn}
R.~Girshick, J.~Donahue, T.~Darrell, and J.~Malik.
\newblock Rich feature hierarchies for accurate object detection and semantic
  segmentation.
\newblock In {\em {CVPR}}, 2014.

\bibitem{SPPnet}
K.~He, X.~Zhang, S.~Ren, and J.~Sun.
\newblock Spatial pyramid pooling in deep convolutional networks for visual
  recognition.
\newblock In {\em ECCV}, 2014.

\bibitem{caffe}
Y.~Jia, E.~Shelhamer, J.~Donahue, S.~Karayev, J.~Long, R.~Girshick,
  S.~Guadarrama, and T.~Darrell.
\newblock Caffe: Convolutional architecture for fast feature embedding.
\newblock {\em arXiv preprint arXiv:1408.5093}, 2014.

\bibitem{krahenbuhl2014geodesic}
P.~Kr{\"a}henb{\"u}hl and V.~Koltun.
\newblock Geodesic object proposals.
\newblock In {\em ECCV}. 2014.

\bibitem{AlexNet}
A.~Krizhevsky, I.~Sutskever, and G.~E. Hinton.
\newblock Imagenet classification with deep convolutional neural networks.
\newblock In {\em NIPS}, 2012.

\bibitem{backprop}
Y.~LeCun, B.~Boser, J.~S. Denker, D.~Henderson, R.~E. Howard, W.~Hubbard, and
  L.~D. Jackel.
\newblock Backpropagation applied to handwritten zip code recognition.
\newblock {\em Neural computation}, 1989.

\bibitem{coco}
T.~Lin, M.~Maire, S.~Belongie, L.~D. Bourdev, R.~B. Girshick, J.~Hays,
  P.~Perona, D.~Ramanan, P.~Doll{\'{a}}r, and C.~L. Zitnick.
\newblock Microsoft {COCO:} common objects in context.
\newblock {\em CoRR}, abs/1405.0312, 2014.

\bibitem{loshchilov2015online}
I.~Loshchilov and F.~Hutter.
\newblock Online batch selection for faster training of neural networks.
\newblock {\em arXiv preprint arXiv:1511.06343}, 2015.

\bibitem{esvm}
T.~Malisiewicz, A.~Gupta, and A.~A. Efros.
\newblock Ensemble of exemplar-svms for object detection and beyond.
\newblock In {\em ICCV}, 2011.

\bibitem{ren2015faster}
S.~Ren, K.~He, R.~Girshick, and J.~Sun.
\newblock Faster {R-CNN}: Towards real-time object detection with region
  proposal networks.
\newblock In {\em Neural Information Processing Systems ({NIPS})}, 2015.

\bibitem{rowley1998neural}
H.~Rowley, S.~Baluja, and T.~Kanade.
\newblock Neural network-based face detection.
\newblock {\em IEEE PAMI}, 1998.

\bibitem{overfeat}
P.~Sermanet, D.~Eigen, X.~Zhang, M.~Mathieu, R.~Fergus, and Y.~LeCun.
\newblock Overfeat: Integrated recognition, localization and detection using
  convolutional networks.
\newblock {\em CoRR}, abs/1312.6229, 2013.

\bibitem{simo2014fracking}
E.~Simo-Serra, E.~Trulls, L.~Ferraz, I.~Kokkinos, and F.~Moreno-Noguer.
\newblock Fracking deep convolutional image descriptors.
\newblock {\em arXiv preprint arXiv:1412.6537}, 2014.

\bibitem{VGG}
K.~Simonyan and A.~Zisserman.
\newblock Very deep convolutional networks for large-scale image recognition.
\newblock {\em CoRR}, abs/1409.1556, 2014.

\bibitem{midlevel}
S.~Singh, A.~Gupta, and A.~A. Efros.
\newblock Unsupervised discovery of mid-level discriminative patches.
\newblock In {\em European Conference on Computer Vision}, 2012.

\bibitem{sungThesis}
K.-K. Sung and T.~Poggio.
\newblock Learning and {E}xample {S}election for {O}bject and {P}attern
  {D}etection.
\newblock In {\em {MIT A.I. Memo No. 1521}}, 1994.

\bibitem{minibatchSVM}
M.~Tak{\'a}{\v{c}}, A.~Bijral, P.~Richt{\'a}rik, and N.~Srebro.
\newblock Mini-batch primal and dual methods for svms.
\newblock {\em arXiv preprint arXiv:1303.2314}, 2013.

\bibitem{Uijlings13}
J.~Uijlings, K.~van~de Sande, T.~Gevers, and A.~Smeulders.
\newblock Selective search for object recognition.
\newblock {\em IJCV}, 2013.

\bibitem{wang2015unsupervised}
X.~Wang and A.~Gupta.
\newblock Unsupervised learning of visual representations using videos.
\newblock In {\em ICCV}, 2015.

\bibitem{ZitnickEdgeBoxes2014}
C.~L. Zitnick and P.~Dollar.
\newblock Edge boxes: Locating object proposals from edges.
\newblock In {\em ECCV}, 2014.

\end{thebibliography}

\end{document}